\documentclass{article}

\PassOptionsToPackage{numbers, compress}{natbib}

\usepackage[preprint]{nips_2018}

\usepackage[utf8]{inputenc} 
\usepackage[T1]{fontenc}    
\usepackage{hyperref}       
\usepackage{url}            
\usepackage{booktabs}       
\usepackage{amsfonts}       
\usepackage{nicefrac}       
\usepackage{microtype}      
\usepackage{amsmath,amssymb,fancyhdr,float,amsthm,listings,multirow,graphicx}
\usepackage{algorithm}
\usepackage{algorithmic}
\usepackage{xcolor}
\usepackage{footnote}

\DeclareMathOperator*{\argmin}{arg\,min}
\DeclareMathOperator*{\argmax}{arg\,max}

\title{Block-wise Partitioning for Extreme Multi-label Classification}

\author{
  Yuefeng Liang \\
  Department of Statistics\\
  University of California, Davis\\
  Davis, CA 95616 \\
  \texttt{frnliang@ucdavis.edu} \\
  \And
  Cho-Jui Hsieh \\
  Department of Statistics\\
  University of California, Davis\\
  Davis, CA 95616 \\
  \texttt{chohsieh@ucdavis.edu} \\  
  \And
  Thomas C.M. Lee \\
  Department of Statistics\\
  University of California, Davis\\
  Davis, CA 95616 \\
  \texttt{tcmlee@ucdavis.edu} \\
}

\begin{document}

\maketitle

\begin{abstract}
 Extreme multi-label classification aims to learn a classifier that annotates an instance with a relevant subset of labels from an extremely large label set. Many existing solutions embed the label matrix to a low-dimensional linear subspace, or examine the relevance of a test instance to every label via a linear scan. In practice, however, those approaches can be computationally exorbitant. To alleviate this drawback, we propose a Block-wise Partitioning (BP) pretreatment that divides all instances into disjoint clusters, to each of which the most frequently tagged label subset is attached. One multi-label classifier is trained on one pair of instance and label clusters, and the label set of a test instance is predicted by first delivering it to the most appropriate instance cluster. Experiments on benchmark multi-label data sets reveal that BP pretreatment significantly reduces prediction time, and retains almost the same level of prediction accuracy.
\end{abstract}

\section{Introduction}

Advances in computing technology enable collection, maintenance, and analysis on extremely large data sets. Domains such as video annotation \cite{snoek2006challenge}, text classification \cite{deng2009imagenet,partalas2015lshtc} and automated tag suggestion \cite{wetzker2008analyzing} generate data sets whose numbers of labels are growing magnificently. Given the abundance of features and labels, supervised learning with hundreds of thousands of labels has attracted attention of machine learning researchers and practitioners in recent years. Automatically assigning a relevant subset of labels from an extremely large label set to an unseen instance defines the goal in extreme multi-label classification \cite{prabhu2014fastxml,bhatia2015sparse}. 

The rapid augmentation of labels leads to numerous computational challenges. In this paper, we focus on one critical issue that limits the performance of existing methods in real applications: {\textbf{prediction time}}. Assume there are $m$ labels, most of the existing methods require $O(m)$ prediction time for each test instance \cite{yu2014large,babbar2017dismec,yen2016pd,yen2017ppdsparse}. For example, the classical one-vs-all approach transforms multi-label problems into multiple binary classification problems. While this approach delivers competitive prediction accuracy on many data sets \cite{babbar2017dismec}, it becomes prohibited when $m$ is extremely large \cite{niculescu2017label}.

To overcome this limitation, we exploit label popularity among all instances in large-scale data sets. It turns out that in many cases, some labels are popular among all instances, while some are frequently tagged only with certain subgroups. Given an association between a label subset and an instance subgroup, prediction performed only from the associated label subset will almost be as accurate as prediction performed from all $m$ labels. This observation leads to our assumption in this paper that the feature and label spaces are so discernible that one can partition instances and labels into different clusters. This partitioning scheme allows us to construct a one-to-one correspondence between each pair of instance and label clusters so that we do not have to visit all $m$ labels for every test instance.

Under our assumption, we show that from feature and label matrices, instance and label clusters can be determined by an alternating update procedure imposed on a discrete optimization problem. If we rearrange labels by label cluster and instances by instance cluster, the permuted label matrix approximates a diagonal block structure, with a majority of ones inside the blocks and zeros elsewhere. Moreover, when an $L_2$ penalty on lengths of label clusters is imposed, the number of labels assigned to each cluster is effectively monitored by the regularization parameter.

Rather than propose a classifier that competes with the existing algorithms, we introduce a Block-wise Partitioning (BP) pretreatment on feature and label matrices to help existing multi-label classifiers achieve faster prediction while retaining prediction accuracy. As for training, one classifier is trained on one pair of instance and label clusters. As for prediction, a test instance is first classified into a proper instance cluster, and then the corresponding classifier is applied on the paired label cluster. We discuss how Precision (P), propensity scored Precision (PSP) and prediction time are impacted by BP pretreatment on large-scale data sets from the Extreme Classification Repository\footnote{http://manikvarma.org/downloads/XC/XMLRepository.html.}. For example, on the Wiki10-31K \cite{zubiaga2012enhancing} data set, while PD-Sparse \cite{yen2016pd} achieves $81.89\%$ at P@1, BP pretreatment accelerates its prediction by 209 times at the expense of $0.33\%$ loss on P@1.

The rest of the paper is organized as follows. We review related work in Section 2 and introduce our partitioning algorithm in Section 3. Experimental results are presented and discussed in Section 4, followed by concluding remarks in Section 5.

\section{Related Work}

Most state-of-the-art methods for extreme multi-label classification can be trichotomized to tree-based, embedding-based and polished one-versus-all approaches. The first branch learns a hierarchical structure of the full label set by recursively dividing feature or label spaces. For instance, FastXML \cite{prabhu2014fastxml} searches for a sparse linear separator to split each node by optimizing an nDCG based loss function. The second branch reduces the effective number of labels under the low-rank assumption. For example, LEML \cite{yu2014large} directly optimizes for the decompression matrices using a regularized least square objective function in a generic empirical risk minimization framework. PD-Sparse \cite{yen2016pd} and DiSMEC \cite{babbar2017dismec} are two methods in the third trend that attract considerable attention in the literature. PD-Sparse makes use of both primal and dual sparsity by margin-maximizing loss with $L_1$ and $L_2$ penalties. DiSMEC revisits one-versus-all paradigm and provides prominent boosts in prediction accuracy and prediction time by explicitly inducing sparsity and doubly parallel training.

Tree-based approaches are well known for their prediction accuracy \cite{prabhu2014fastxml}, and embedding-based approaches are also popular as they are straightforward and can handle label correlations \cite{yu2014large,bhatia2015sparse}. However, implementation on large data sets has been a subject of extensive debate, and one of the most common remarks is on prediction time \cite{si2017gradient,niculescu2017label}. Recently, various methods have been proposed to ameliorate the burden. For instance, PPD-Sparse \cite{yen2017ppdsparse} adapts the parallelizability and small memory footprint of the one-versus-all technique from DiSMEC, and the sub-linear complexity of primal-dual sparse method from PD-Sparse.

In contrast to the above methods, we develop a pretreatment for those algorithms. The following three approaches are relevant to ours. Label Partitioning for Sub-linear Ranking (LPSR) follows a two-step approach. At the training stage, it clusters the training instances, and then assigns a fixed number of potential labels to each cluster. At the testing stage, a test instance is first put in one of the clusters, and its labels are predicted only from the labels attached to that cluster \cite{weston2013label}. While LPSR and our proposal stand on the same testing stage, three differences can be observed at the training stage. First, the optimized number of clusters in our method is selected by the algorithm, whereas the number in LPSR is predefined. Second, cardinality of our label clusters may vary across clusters, while the number of potential labels in each cluster in LPSR is fixed. Moreover, our instance and label clusters are updated via optimization, whereas assignment of training instances in LPSR solely depends on feature matrix. Consequently, our method permits more flexibility in cluster structure.

The second method, Clustering Based Multi-Label Classification (CBMLC), groups the training instances into a user-specified number of clusters, and trains a multi-label classification model for each cluster \cite{nasierding2009clustering}. The testing stage is identical to that of LPSR. There are several differences between CBMLC and our approach. First, our goal is to minimize the prediction time while CBMLC focuses on reducing the training time. Second, we formally form a clustering objective to achieve better prediction time using both features and labels, and expedite prediction by reducing label size, whereas CBMLC only takes features into consideration during the clustering step, and label size remains the same. Furthermore, our number of clusters is not user-specified.

Another related approach is to pre-select a small fraction of candidate labels via label filters (LF) \cite{niculescu2017label} before the base classifier is applied. As the number of filters increases, a larger proportion of labels will be filtered out, and thus prediction time decreases without significant impact on prediction performance. Although LF shares the same goal as we do -- to speed up existing classifiers at prediction time by reducing the number of candidate labels, infrequent, unnoticeable but valuable tail labels will be filtered out by LF; those labels can be captured and stay in one of our label clusters, given that they are closely related to a group of training instances. What is more, LF tends to select the same candidate labels for all instances, while our method assigns customized label subsets to different instance clusters. As a result, our method reflects more natural characteristics of features and labels without discarding rare but rewarding signals.

\section{Proposed Algorithm}

\subsection{Problem Formulation} 

To formulate our partitioning problem, we adopt the following notations: lower-case letters such as $x$ and $y$ denote elements of sets; upper-case letters such as $C$ and $L$ denote maps; bold lower-case letters such as $\bf{x}$ and $\bf{y}$ denote vectors; bold upper-case letters such as $\bf{X}$ and $\bf{Y}$ denote matrices; calligraphic letters such as $\mathcal{X}$ and $\mathcal{L}$ denote sets.

Let $\mathbf{x}_i=(x_{i1},x_{i2},\ldots,x_{id})\in\mathbb{R}^d$, $i=1,\cdots,n$, be a $d$-dimensional input feature vector that forms an instance, and $\mathbf{y}_i=(y_{i1},y_{i2},\ldots,y_{im})\in\{0,1\}^m$,\ $i=1,\cdots,n$, be the corresponding $m$-dimensional label vector. $y_{ij}=1$ if the $i^\text{th}$ data point is tagged with the $j^\text{th}$ label. Let $\{(\mathbf{x}_i,\mathbf{y}_i)\}^n_{i=1}$ be the training data set. $\mathbf{X}=[\mathbf{x}_1,\ldots,\mathbf{x}_n]^\top$ is the $n\times d$ feature matrix, and $\bf{Y}=[\mathbf{y}_1,\ldots,\mathbf{y}_n]^\top$ is the $n\times m$ label matrix. We are interested in simultaneously partitioning instances and labels into $q$ $(1\leq q\leq \min(m,n))$ instance and label clusters. Write the $q$ instance and label clusters as $\{\mathcal{X}_1,\mathcal{X}_2,\ldots,\mathcal{X}_q\}$ and $\{\mathcal{L}_1,\mathcal{L}_2,\ldots,\mathcal{L}_q\}$ respectively, where $\mathcal{X}_l\subseteq\{1,\ldots,n\}$ and $\mathcal{L}_l\subseteq\{1,\ldots,m\}$ for $l=1,\ldots, q$. We define a one-to-one deterministic mapping $L$ from instance clusters to label clusters such that 
\begin{equation*}
L(\mathcal{X}_l)=\mathcal{L}_l \text{ for all }l. 
\end{equation*}
Therefore, our goal is to learn a map
\begin{equation*}
C: \mathbb{R}^d\longrightarrow\{\mathcal{X}^*_1,\mathcal{X}^*_2,\ldots,\mathcal{X}^*_q\}
\end{equation*}
that classifies instances, and an optimal set of label clusters $\{\mathcal{L}^*_1,\mathcal{L}^*_2,\ldots,\mathcal{L}^*_q\}$. 

The partitioned label matrix approximates a block diagonal matrix upon row and column permutations. The diagonal blocks are matrices of any size with a majority of ones, and the off-diagonal elements are mostly zeros. Instance clusters are deemed to be disjoint, and overlapping is allowed between any two label clusters, which means there can be overlap between $\mathcal{L}^*_l$ and $\mathcal{L}^*_{l'}$ for any $(l, l')$ pair.  

\begin{figure}[ht]
  \vskip -0.1in
  \centering
  \centerline{\includegraphics[width=.4\columnwidth]{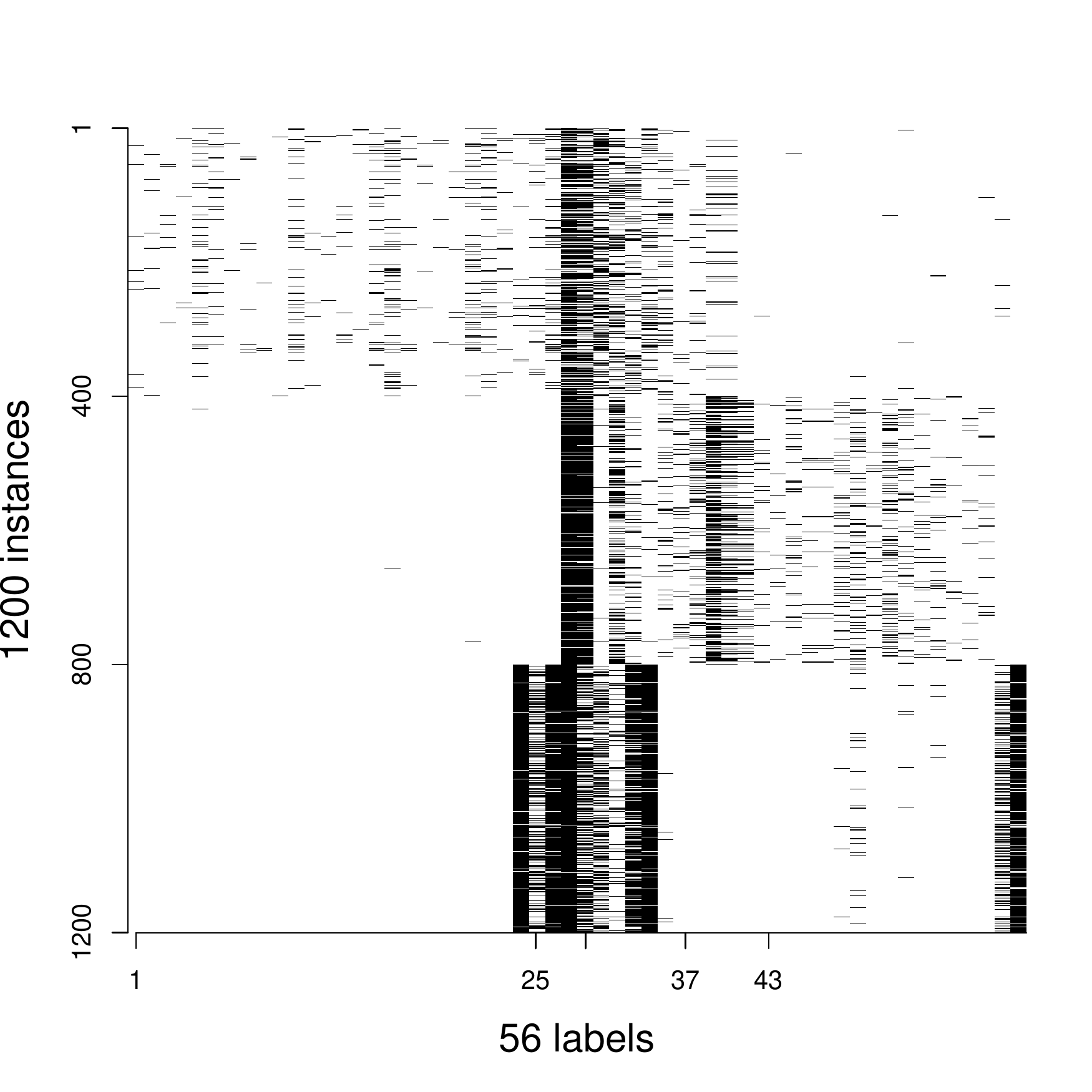}}
  \caption{Truncated and permuted training label matrix of Mediamill. Black pixels are tagged labels.}
  \label{mediamill}
  \vskip -0.1in
\end{figure}

Allowing overlapping label clusters is crucial for achieving good performance on extreme classification problems. As we state in Section 1, some popular labels are often assigned to many samples, and our clustering approach is able to capture this information by assigning those popular labels to many clusters. Figure~\ref{mediamill} illustrates the approximate block diagonal structure on small-scale data set Mediamill \cite{snoek2006challenge}. For illustration purposes, $q$ is set to $3$, $56$ out of $m=101$ labels are selected in label clusters, and only the first $400$ train instances from each instance cluster are displayed. It can be visualized that three labels (from the $34^{th}$ to the $36^{th}$) are allocated to all label clusters. If these popular labels were assigned to only one label cluster, then the other two would fail to capture the strong signals and thus suffered from an undesired loss in predictive power. 

After we find the optimal partitions, we learn a classifier from each pair of $\mathcal{X}_l$ and $\mathcal{Y}_l$. Finally, given a test instance $\hat{\mathbf{x}}$, we first classify it into one of $\{\mathcal{X}^*_1,\mathcal{X}^*_2,\ldots,\mathcal{X}^*_q\}$ by conducting a multinomial logistic regression on the feature space, and then predict its most relevant labels $\hat{\mathbf{y}}$ from its label cluster. 

\subsection{Block-wise Partitioning (BP)}

As we mention in Section 3.1, our goal is to learn a function $C$ and a set of label clusters $\{\mathcal{L}^*_1,\mathcal{L}^*_2,\ldots,\mathcal{L}^*_q\}$ for any given $q$. Let $\Omega_q=(C, \{\mathcal{L}_1,\mathcal{L}_2,\ldots,\mathcal{L}_q\})$. We wish to find the optimal $\Omega_q^*$ which minimizes the objective function 
\begin{equation}
f(\Omega_q)=-\sum^n_{i=1}\sum^m_{j=1}\mathbf{1}[j\in (L\circ C)(\mathbf{x}_i), y_{ij}=1]+\lambda\sum^q_{l=1}|\mathcal{L}_l|^2,
\label{eq:obj}
\end{equation}
where $\lambda$ is a regularization parameter. The intuition behind $f(\Omega_q)$ is to search for an equilibrium between the number of ones captured by the diagonal blocks and the total size of label clusters. We first observe that this optimization problem is a significant challenge, as the objective function is neither convex nor differentiable. Furthermore, our goal is to perform optimization for data sets where $d, m$, and $n$ are substantially large. Nevertheless, when either one of the two sets of clusters is fixed, the objective becomes a convex function. To this end, we divide the optimization problem into two phases, and implement alternating minimization. 

To start with, we specify $q$, the initial number of paired clusters, and set $t=0$. The selection criteria of $q$ will be discussed in Section 3.3. Given a chosen $q$, we initialize $C$ by applying the sparse $k$-means clustering on the entire training feature matrix. In other words, we initialize $\{\mathcal{X}^{(0)}_1,\mathcal{X}^{(0)}_2,\ldots,\mathcal{X}^{(0)}_q\}$ by features.

{\textbf{Label Clusters Selection:}} From $t=1$, we fix $\{\mathcal{X}^{(t-1)}_1,\mathcal{X}^{(t-1)}_2,\ldots,\mathcal{X}^{(t-1)}_q\}$ and update $\{\mathcal{L}^{(t)}_1,\mathcal{L}^{(t)}_2,\ldots,\mathcal{L}^{(t)}_q\}$.
This is a discrete optimization problem and a naive implementation will need to check the objective function~\eqref{eq:obj} for every possible set of labels. Instead, we find the optimal set can be efficiently computed using the following procedure. 

\begin{figure*}[!]
  \centering
  \vskip -0.1in
  \centerline{\includegraphics[width=.4\columnwidth]{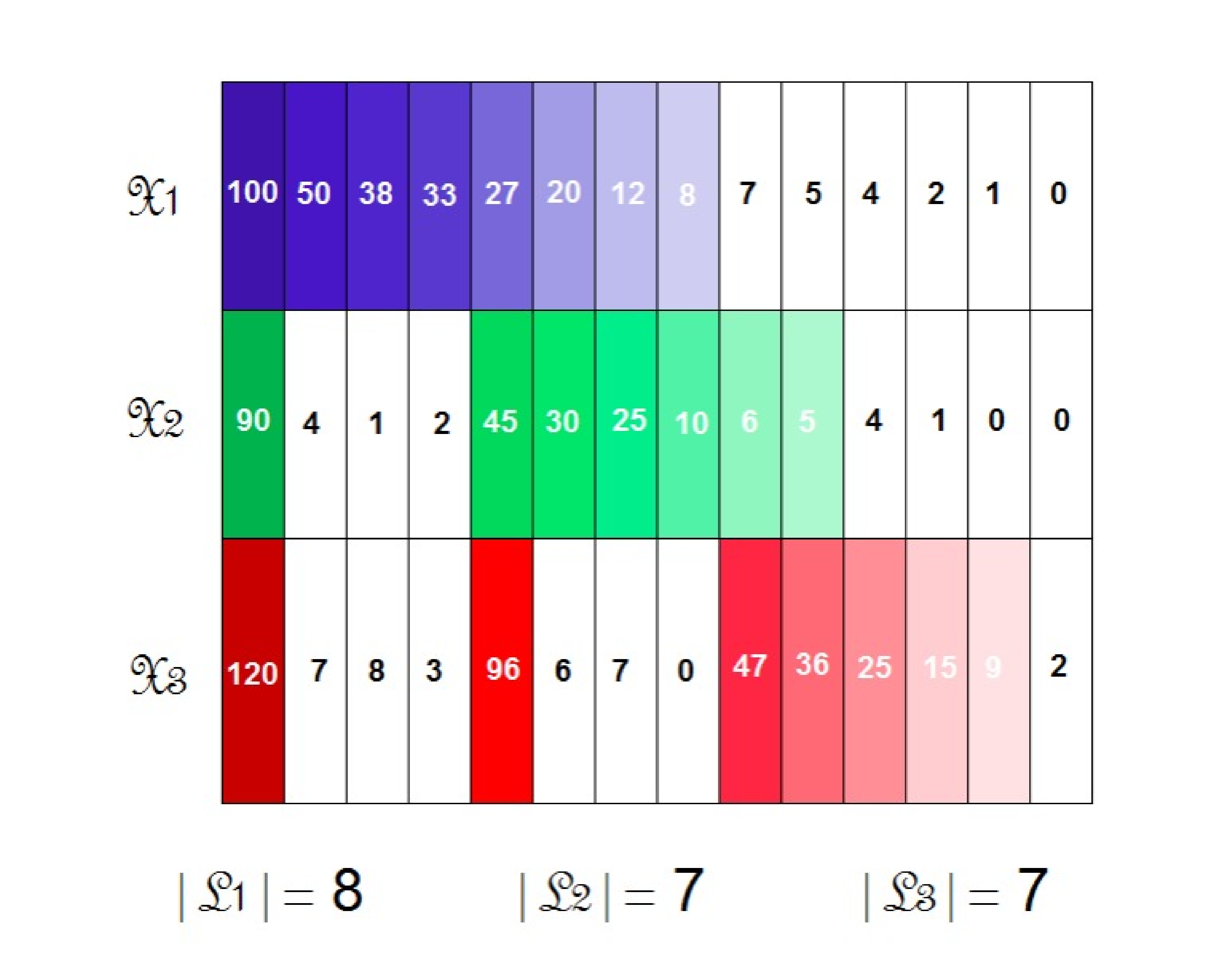}
  \includegraphics[width=.4\columnwidth]{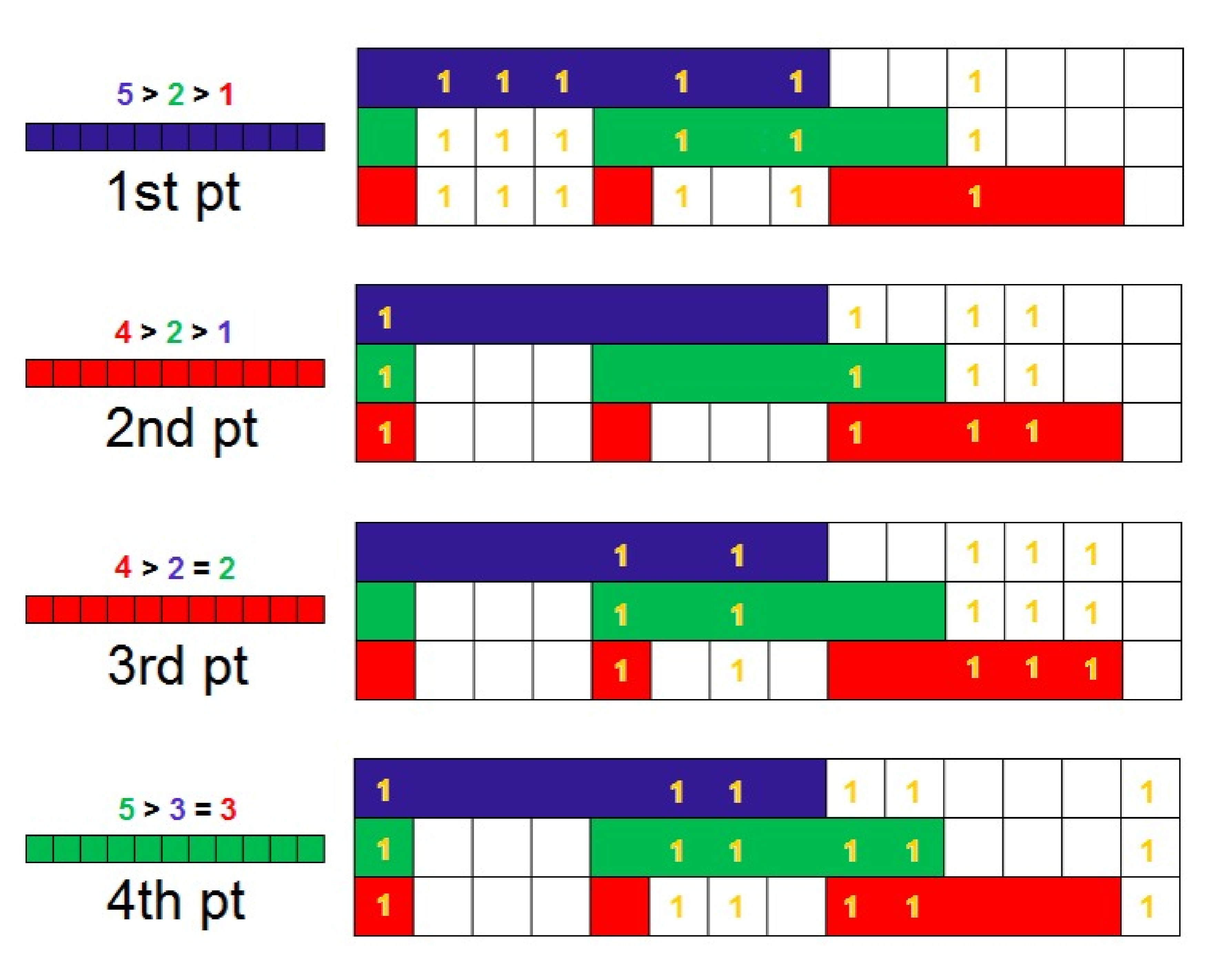}}
  \caption{An example of label clusters selection (left) and instance clusters selection (right). For label clusters selection, lighter colors represent lower column sums. Best viewed in color.}
  \label{block}
\end{figure*}

Given an instance cluster $\mathcal{X}_l$, we first calculate column sums in the label matrix for all $m$ labels, and then sort the $m$ column sums in descending order. 
By the definition of objective function, we know if one label is included in $\mathcal{L}_l$, then all the labels with larger column sum should also be included in $\mathcal{L}_l$ (otherwise exchanging two lables will decrease objective function). Thus we add the sorted $m$ labels one by one to the label set $\mathcal{L}_l$, and each time check the objective function
\begin{equation}
-\sum_{i\in\mathcal{X}_l}\sum^J_{j=1}\mathbf{1}[j\in (L\circ C)(\mathbf{x}_i)]\cdot\mathbf{1}[y_{ij}=1]+\lambda J^2
\label{eq:obj2}
\end{equation}
until adding one additional label will increase \eqref{eq:obj2}. It is obvious that further increasing $J$ will lead to larger objective function value, thus assigning the first $J$ labels to $\mathcal{L}_l$ will be optimal. Figure~\ref{block} (left) visualizes Algorithm~\ref{labelselect} via a simple example with $q=3$ and $m=14$. Different colors distinguish different clusters. The top-left number $100$, for instance, is the column sum of the first column, i.e., label 1, in the training label matrix for all samples in $\mathcal{X}_1$. As the color turns lighter from left to right, the column sums decrease, and the algorithm selects the first eight labels in the corresponding label cluster $\mathcal{L}_1$.

In this procedure, computing the count of each label in each instance cluster costs $O(\text{nnz}({\bf{Y}}))$ time in total and sorting for all the clusters costs $O(q m\log (m))$ time. Once the label counts are computed and sorted, every step of checking~\eqref{eq:obj2} only costs $O(1)$ time, so the overall time complexity for this step is $O(\text{nnz}({\bf{Y}})+qm\log(m))$.

{\textbf{Instance Clusters Selection:}} Now we fix $\{\mathcal{L}^{(t)}_1,\mathcal{L}^{(t)}_2,\ldots,\mathcal{L}^{(t)}_q\}$ and update $\{\mathcal{X}^{(t)}_1,\mathcal{X}^{(t)}_2,\ldots,\mathcal{X}^{(t)}_q\}$. We take the following steps for every instance. We first assign an instance to every one of the $q$ instance clusters, and then count the number of attached labels that belong to the corresponding label cluster. If the $I^\text{th}$ assignment yields the greatest count, then we put the instance in $\mathcal{X}_I$. Figure~\ref{block} (right) envisions Algorithm~\ref{instanceselect} for the first four sample points under the same framework as Figure~\ref{block} (left). The $1$’s in yellow are the ground truth labels in training label matrix. For the first sample, there are five $1$’s in blue, two in green, and one in red. At this step, we fix label clusters and update instance clusters. If we place the first sample into $\mathcal{X}_1$, then five ground truth labels will be captured in $\mathcal{L}_1$. The same idea follows for green and red. Since placing it into $\mathcal{X}_1$ yields the highest number of ground truth labels to be captured, we put the first sample into $\mathcal{X}_1$, and mark it in blue. 

In this procedure, computing the count of each instance in each label cluster costs $O(\text{nnz}({\bf{Y}}))$ time in total, and comparing the counts of $q$ assignments costs $O(q)$ time for each instance, so the overall time complexity for this step is $O(\text{nnz}({\bf{Y}})+nq)$.

We iterate the above two selection procedures until the objective reaches a local minimum. Once the difference between $f(\Omega^{(t-1)}_q)$ and $f(\Omega^{(t)}_q)$ is less than $10^{-5}$, we stop at time $t$, and report $\{\mathcal{X}^{(t)}_1,\mathcal{X}^{(t)}_2,\ldots,\mathcal{X}^{(t)}_q\}$ and $\{\mathcal{L}^{(t)}_1,\mathcal{L}^{(t)}_2,\ldots,\mathcal{L}^{(t)}_q\}$ 
as the optimal set of instance and label clusters, respectively. 

As the time complexities of label clusters selection and instance clusters selection are $O(\text{nnz}({\bf{Y}})+qm\log(m))$ and $O(\text{nnz}({\bf{Y}})+nq)$ respectively, the total time complexity of this alternating minimization procedure is $O(\text{nnz}({\bf{Y}})+qm\log(m)+nq)$.\\

\begin{minipage}[t]{8cm}
  \vskip -0.1in  
  \begin{algorithm}[H]
    \caption{Label Clusters Selection}
    \label{labelselect}
    \begin{algorithmic}
    \STATE {\bfseries Input:} $\{\mathcal{X}^{(t-1)}_l\}_{l=1}^q$
    \STATE {\bfseries Output:} $\{\mathcal{L}^{(t)}_l\}_{l=1}^q$
    \FOR{$l=1$ {\bfseries to} $q$}
    \STATE $P^{(t-1)}_{lj}=\sum_{i\in\mathcal{X}_l^{(t-1)}}\mathbf{1}[y_{ij}=1]$ for all $j$
    \STATE Sort $\{P^{(t-1)}_{lj'}\}_{j'=1}^m$ in descending order
    \STATE Find $J^*=\argmin_{J}-\sum_{j'=1}^{J}P^{(t-1)}_{lj'}+\lambda J^2$ using a linear scan
    \STATE Assign $\{1,\ldots,J^*\}$ to $\mathcal{L}_l^{(t)}$
    \ENDFOR
    \end{algorithmic}
  \end{algorithm}
  \vskip -0.2in 
\end{minipage}%
\begin{minipage}[t]{8cm}
  \vskip -0.1in 
  \begin{algorithm}[H]
    \caption{Instance Clusters Selection}
    \label{instanceselect}
    \begin{algorithmic}
    \STATE {\bfseries Input:} $\{\mathcal{L}^{(t)}_l\}_{l=1}^q$
    \STATE {\bfseries Output:} $\{\mathcal{X}^{(t)}_l\}_{l=1}^q$
    \FOR{$i=1$ {\bfseries to} $n$}
    \STATE $Q^{(t)}_{il}=\sum_{j\in\mathcal{L}^{(t)}_l}\mathbf{1}[y_{ij}=1]$ for all $l$
    \STATE Find $I=\argmax_lQ^{(t)}_{il}$
    \STATE Assign $i$ to $\mathcal{X}^{(t)}_{I}$
    \ENDFOR
    \end{algorithmic}
  \end{algorithm}
  \vskip -0.2in 
\end{minipage}

\subsection{Optimal Parameters}

The regularization parameter $\lambda$ monitors the tradeoff between prediction speed and prediction accuracy. The optimal $\lambda$ depends on one's preference on prediction efficiency and tolerance of accuracy loss. Conditional on the tolerance level, we are able to apply cross-validation to select the optimal $\lambda$. For each validation set, we record an interval of $\lambda$'s that keeps the loss of prediction accuracy within the tolerance level. The optimal range of $\lambda$'s is taken as the intersection of all intervals. In real data experiments, we conduct $5$-fold cross-validation, and set the tolerance level to $2\%$.

Note that our algorithm assumes that the numbers of label clusters and instance clusters are equal to $q$, but in practice, the ``true'' number of paired clusters is unknown. Nevertheless, our algorithm automatically searches for the optimal $q$ within a search space. On the one hand, for a fixed $\lambda$, if the diagonal blocks in training label matrix capture as many ones as possible, then important labels are likely to stay in at least one of the label clusters, and thus predictive power is unaffected. On the other hand, if $q$ is too large, some of the paired clusters may be empty. Hence, there exists a unique $q$ such that the maximal proportion of ones in the full training label matrix is attained, and all $q$ paired clusters are non-empty. For instance, in the large-scale data set AmazonCat-13K \cite{mcauley2013hidden}, when $\lambda=1$, as $q$ increases from $2$ to $5$, the proportion of $1$'s being captured in clusters increases from $80\%$ to $84\%$. If we take $q=6$, one pair of instance and label clusters shrinks to null. Therefore, our algorithm selects $q=5$ for AmazonCat-13K.

\subsection{Prediction}

To predict labels of test instances, we allocate all instances into proper instance clusters based on their features. This task is a supervised learning problem, which is handled by the linear classification package LIBLINEAR \cite{fan2008liblinear}. We use L2-regularized linear logistic regression in LIBLINEAR, with the default one-versus-all approach for multi-class classification. The training is efficient since LIBLINEAR only takes $O(\text{nnz}({\bf{X}})q)$ time per epoch. The final step of the whole process is to apply the trained classifiers on the assigned testing instance cluster and corresponding label cluster. The prediction only requires $q$ inner products, and $q$ is much smaller than the original label size. Hence, prediction time complexity is reduced from $O(m)$ to $O(|\mathcal{L}^*_l|)$. To show efficiency of LIBLINEAR, we report the train and prediction time, together with the number of instance clusters in all data sets on which we experiment in the Appendix.

\section{Experiments}

\begin{table*}[ht]
\caption{Data set statistics for extreme multi-label classification problems. ASpL and ALpS represent average sample per label and average labels per sample respectively.}
\label{statistics}
\vskip 0.15in
\begin{center}
\begin{small}
\begin{tabular}{crrrrrr} 
 \hline
 Data & \# Training & \# Testing & \# Features & \# Labels & ASpL & ALpS \\
 \hline
 AmazonCat-13K & 1,186,239 & 306,782 & 203,882 & 13,330 & 448.57 & 5.04 \\
 Wiki10-31K & 14,146 & 6,616 & 101,938 & 30,938 & 8.52 & 18.64 \\
 Delicious-200K & 196,606 & 100,095 & 782,585 & 205,443 & 72.29 & 75.54 \\
 WikiLSHTC-325K & 1,778,351 & 587,084 & 1,617,899 & 325,056 & 17.46 & 3.19 \\
 Amazon-670K & 490,449 & 153,025 & 135,909 & 670,091 & 3.99 & 5.45 \\
 \hline
\end{tabular}
\end{small}
\end{center}
\vskip -0.1in
\end{table*}

We evaluate the performance of BP pretreatment on combinations of data sets and classifiers with respect to prediction accuracy and prediction time.

\textbf{Data:} We conduct experiments on five large-scale data sets from the Extreme Classification Repository. Table~\ref{statistics} shows the associated details. We use the same train-test splits as the Repository.

\textbf{Classifiers:} We test BP pretreatment on the first three extreme multi-label classifiers and compare the results with those of the last two approaches.
\begin{itemize}
\item LEML \cite{yu2014large} is a low-rank Empirical-Risk-Minimization solver. We implement BP with this technique because its prediction time complexity is linear to $m$.
\item PD-Sparse \cite{yen2016pd} uses $L_1$ regularization along with multi-class loss. Since the intermediary weight vectors need to be stored in size linear to $m$, we implement BP with this algorithm.
\item DiSMEC \cite{babbar2017dismec} is a distributed framework based on one-versus-all linear classifiers with explicit model size controlled by pruning small weights. 
\item CBMLC \cite{nasierding2009clustering} comprises a clustering algorithm and a multi-label classification algorithm. We instantiate the clustering component using the $k$-means algorithm and implement the classification component using the above three classifiers.
\item LPSR-NB \cite{weston2013label} learns a hierarchy of labels as well as a classifier for the entire label set. Since computational cost for each task is high, training Naive Bayes classifier as the base classifier is the only possible strategy to large-scale data sets. Results of this approach is taken from Table 1(a) in the SLEEC paper \cite{bhatia2015sparse}, Table 2 in the DiSMEC paper \cite{babbar2017dismec} and the Repository.
\end{itemize}

\textbf{Parameter Setting:} For LEML, we set the rank to $500$ and the number of iterations to $5$ for an early stop\footnote{These choices are adopted from the Extreme Classification Repository.}. All other parameters are set as default. For PD-Sparse and DiSMEC, we use the default parameters provided by the authors.

\textbf{Evaluation Metrics:} Precision@$k$ (P@$k$) counts the fraction of correct predictions in the top $k$ scoring labels in $\hat{\mathbf{y}}$ (this is a standard evaluation on Extreme Repository) \cite{hsu2009multi,agrawal2013multi}. For selected classifiers, we also present propensity scored Precision@$k$ (PSP@$k$), as propensity score helps in making metrics unbiased \cite{jain2016extreme}. To evaluate improvement in prediction efficiency, we consider speed-up, a measure with regard to number of vector multiplications \cite{niculescu2017label}.

\begin{table*}[!]
\caption{Elapsed time (in seconds) for speed-driven LEML and DiSMEC with BP pretreatment.}
\label{Time}
\begin{center}
\begin{small}
\begin{tabular}{llrrrrr} 
 \hline
 Method & Procedure & A.-13K & W.-31K & D.-200K & W.-325K & A.-670K \\
 \hline
 BP LEML & Data Partitioning & 2102 & 107 & 1675 & 7841 & 3049 \\
 BP LEML & Training & 17626 & 1239 & 4581 & 8800 & 5753 \\
 LEML & Original Training & 28200 & 4448 & 18599 & 43473 & 9802 \\
 \hline
 BP DiSMEC & Data Partitioning & 220 & 6 & 118 & 89613 & 35735 \\
 BP DiSMEC & Training & 6587 & 541 & 4134 & 95243 & 49348 \\
 DiSMEC & Original Training & 12459 & 2339 & 39439 & 300655 & 193842 \\
 \hline
\end{tabular}
\end{small}
\end{center}
\vskip -0.1in
\end{table*}

\begin{figure*}[!]
\begin{center}
\centerline{\includegraphics[width=.33\columnwidth]{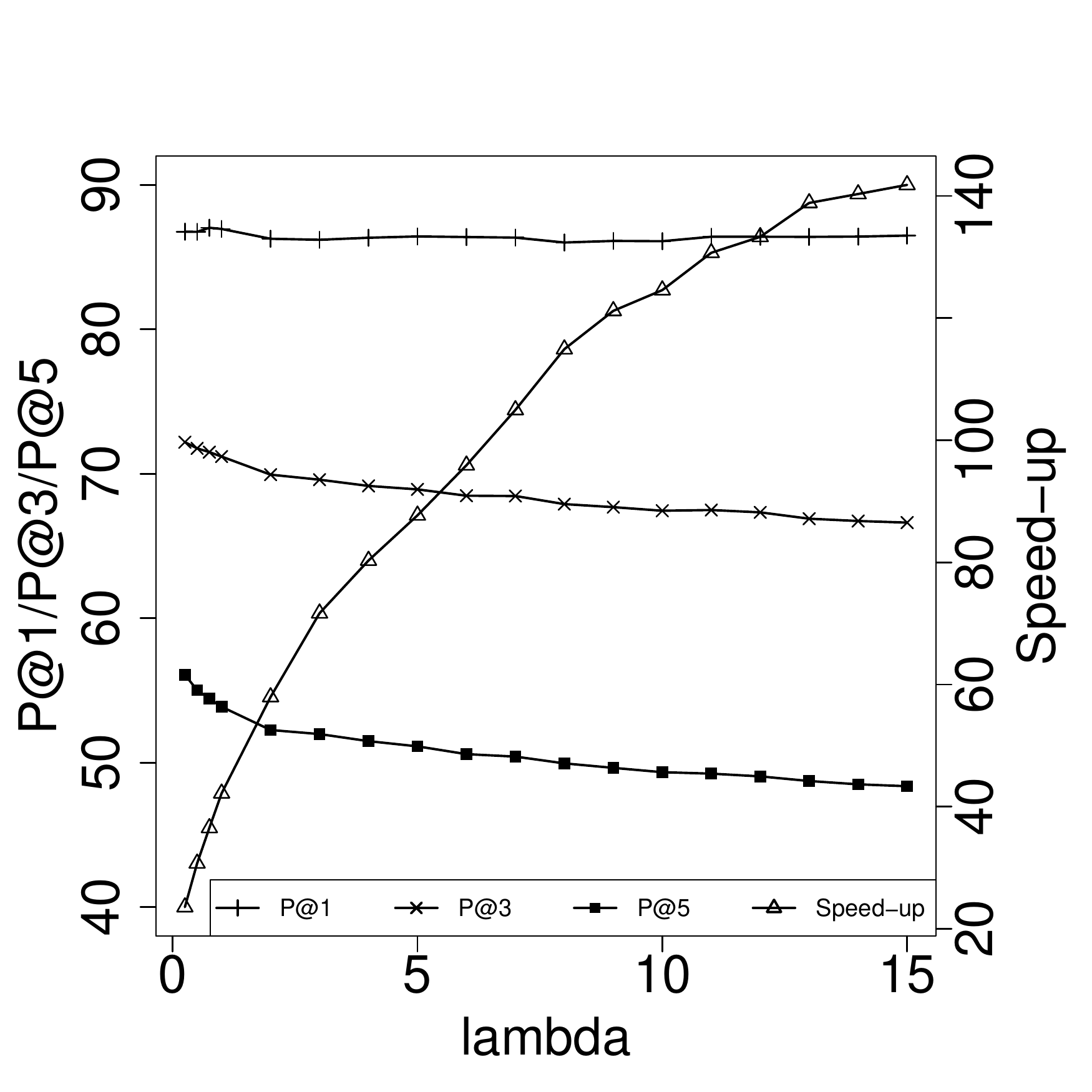}
\includegraphics[width=.33\columnwidth]{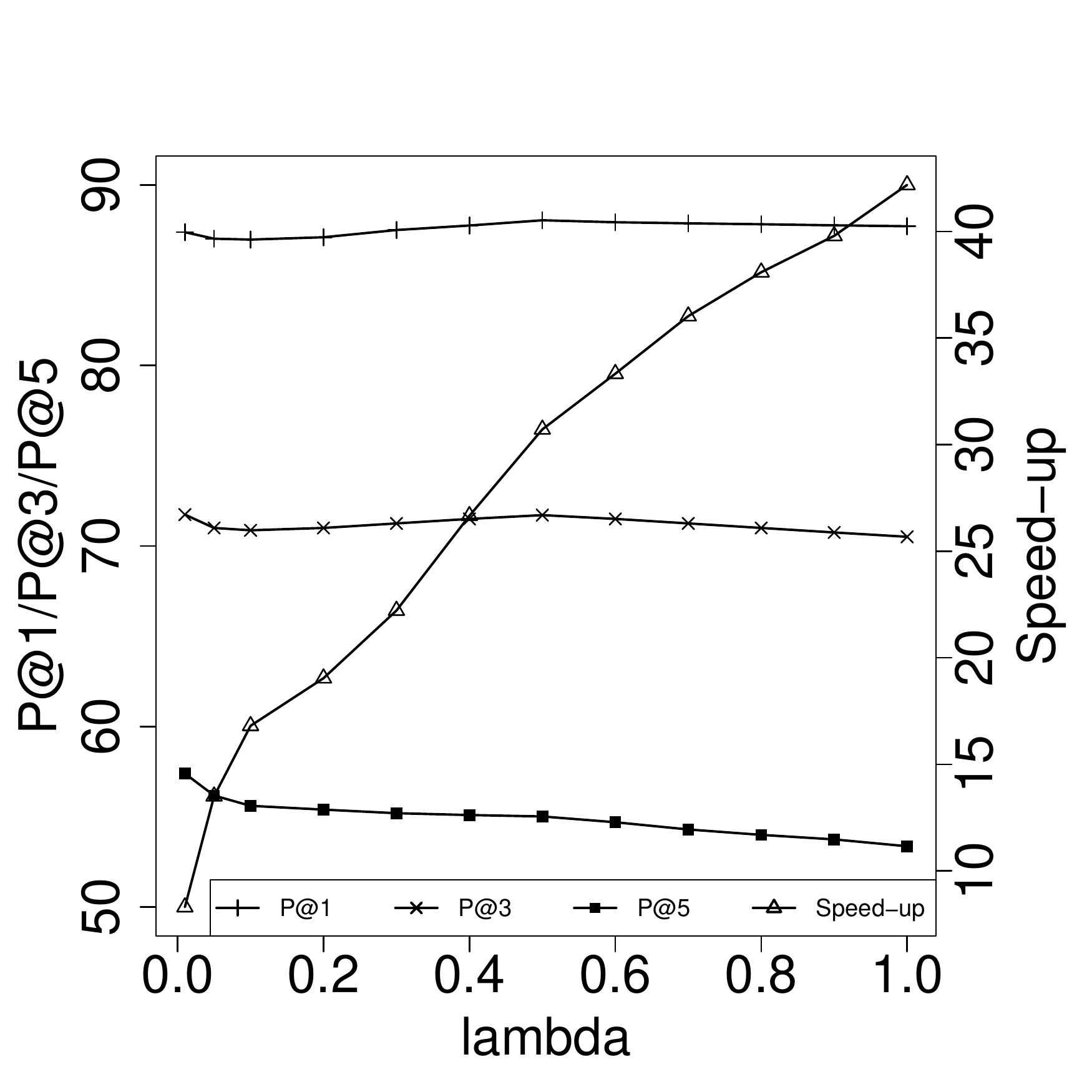}
\includegraphics[width=.33\columnwidth]{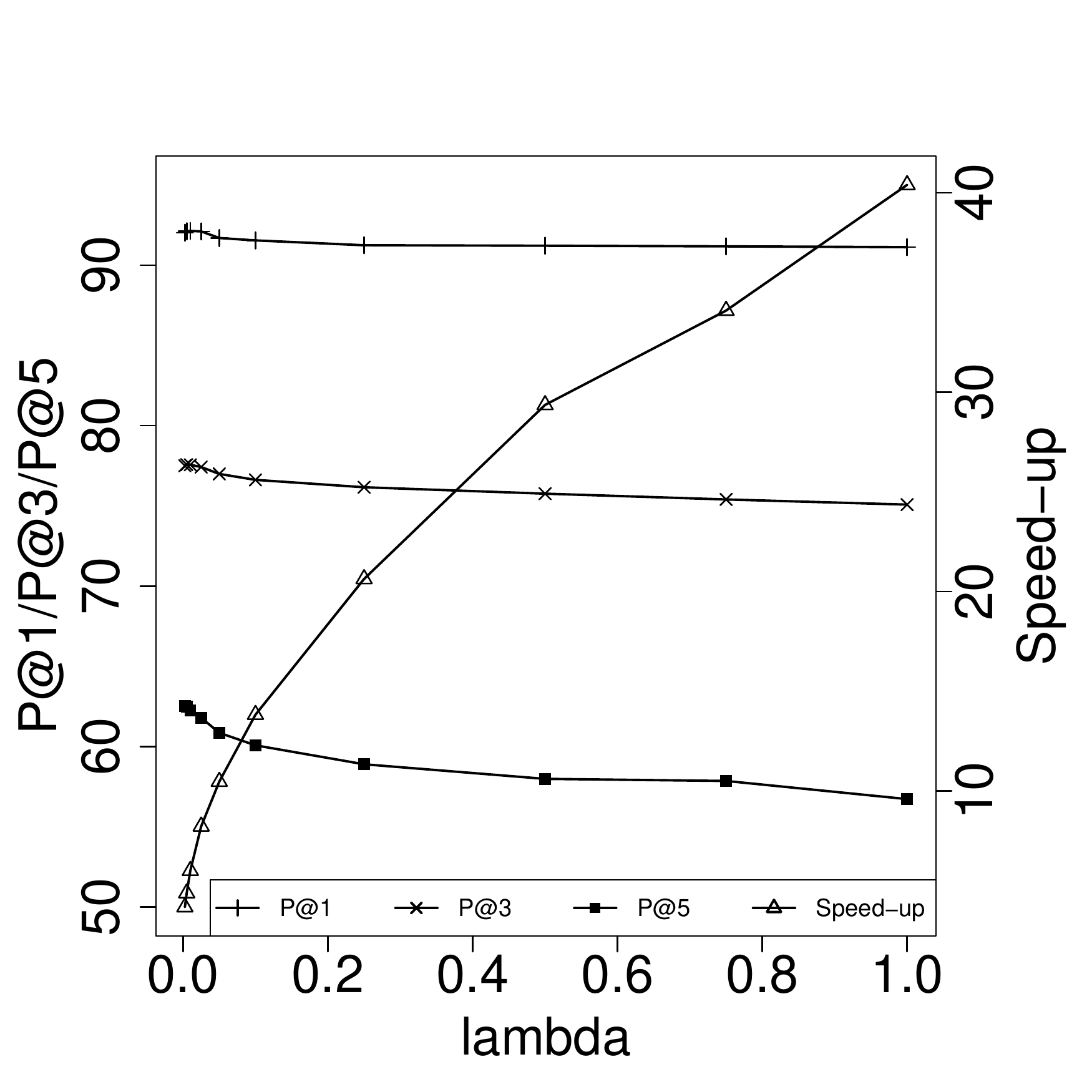}}
\centerline{\includegraphics[width=.33\columnwidth]{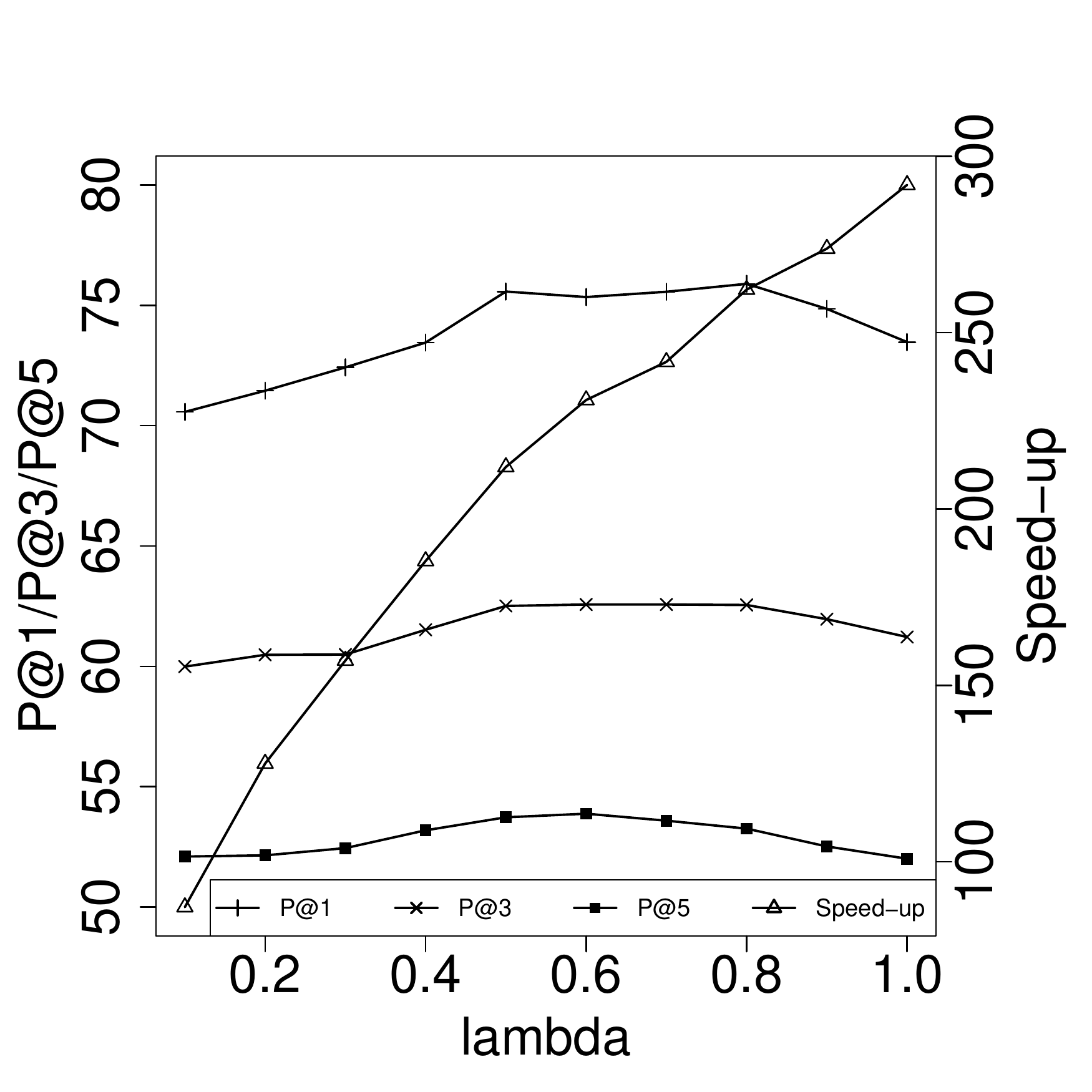}
\includegraphics[width=.33\columnwidth]{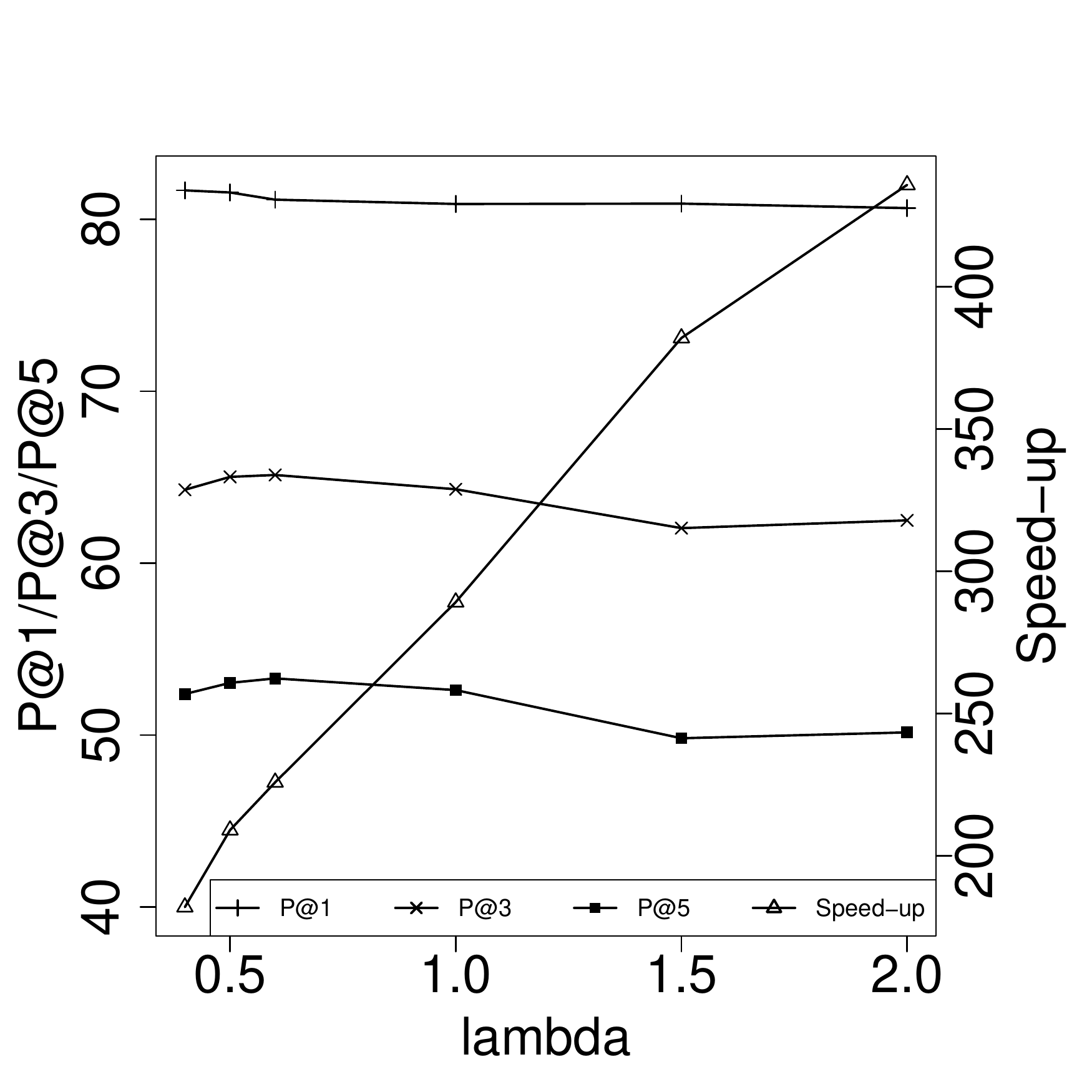}
\includegraphics[width=.33\columnwidth]{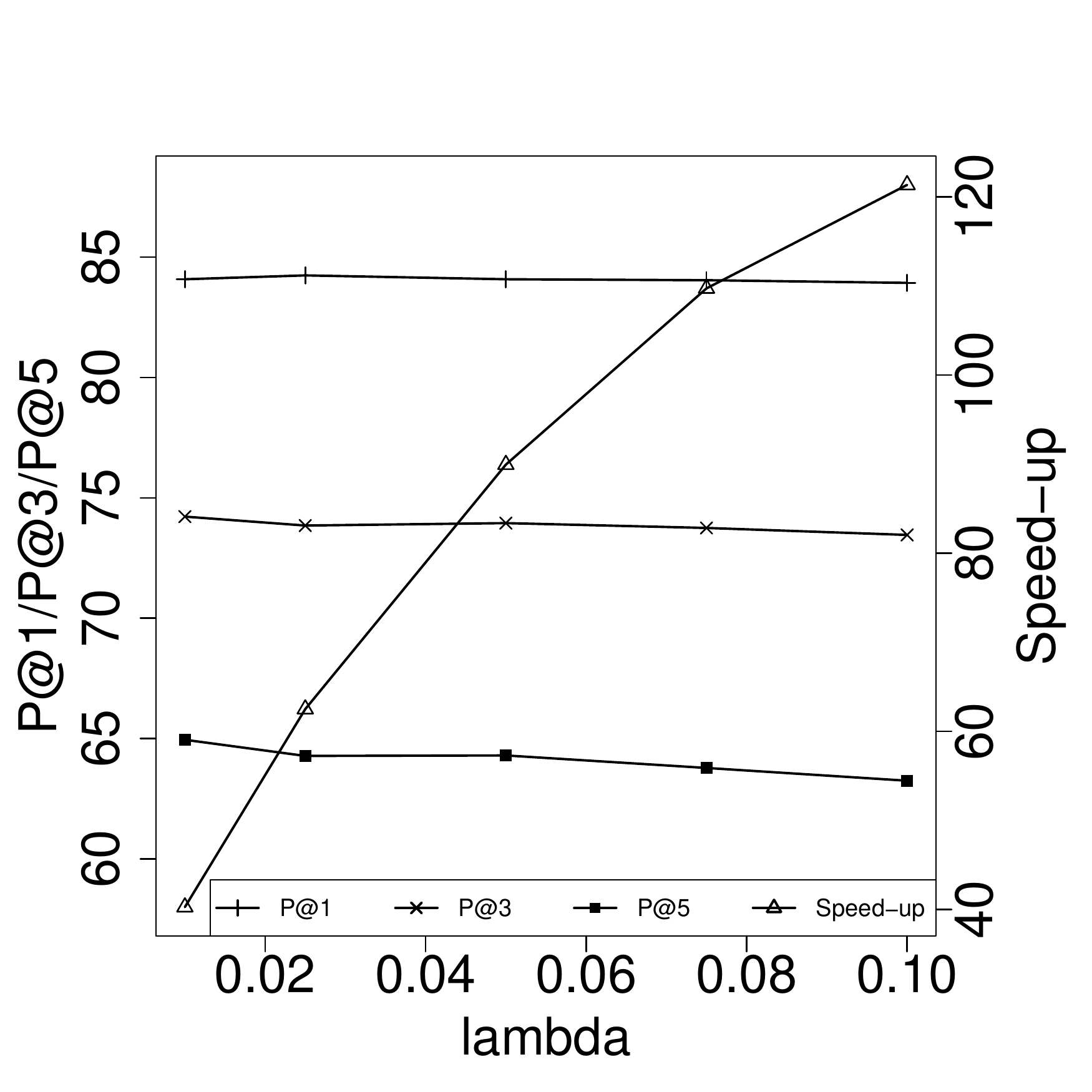}}
\centerline{\includegraphics[width=.33\columnwidth]{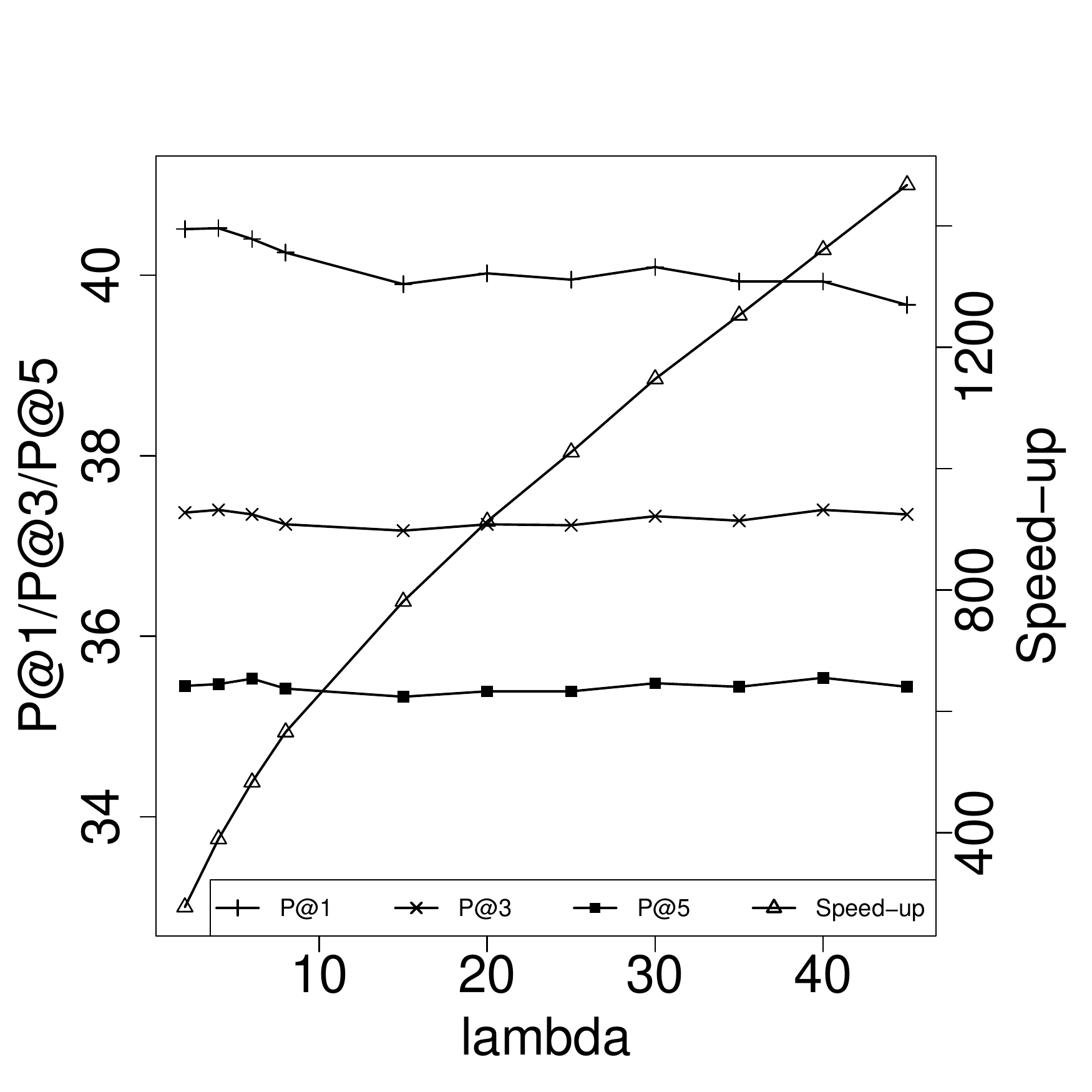}
\includegraphics[width=.33\columnwidth]{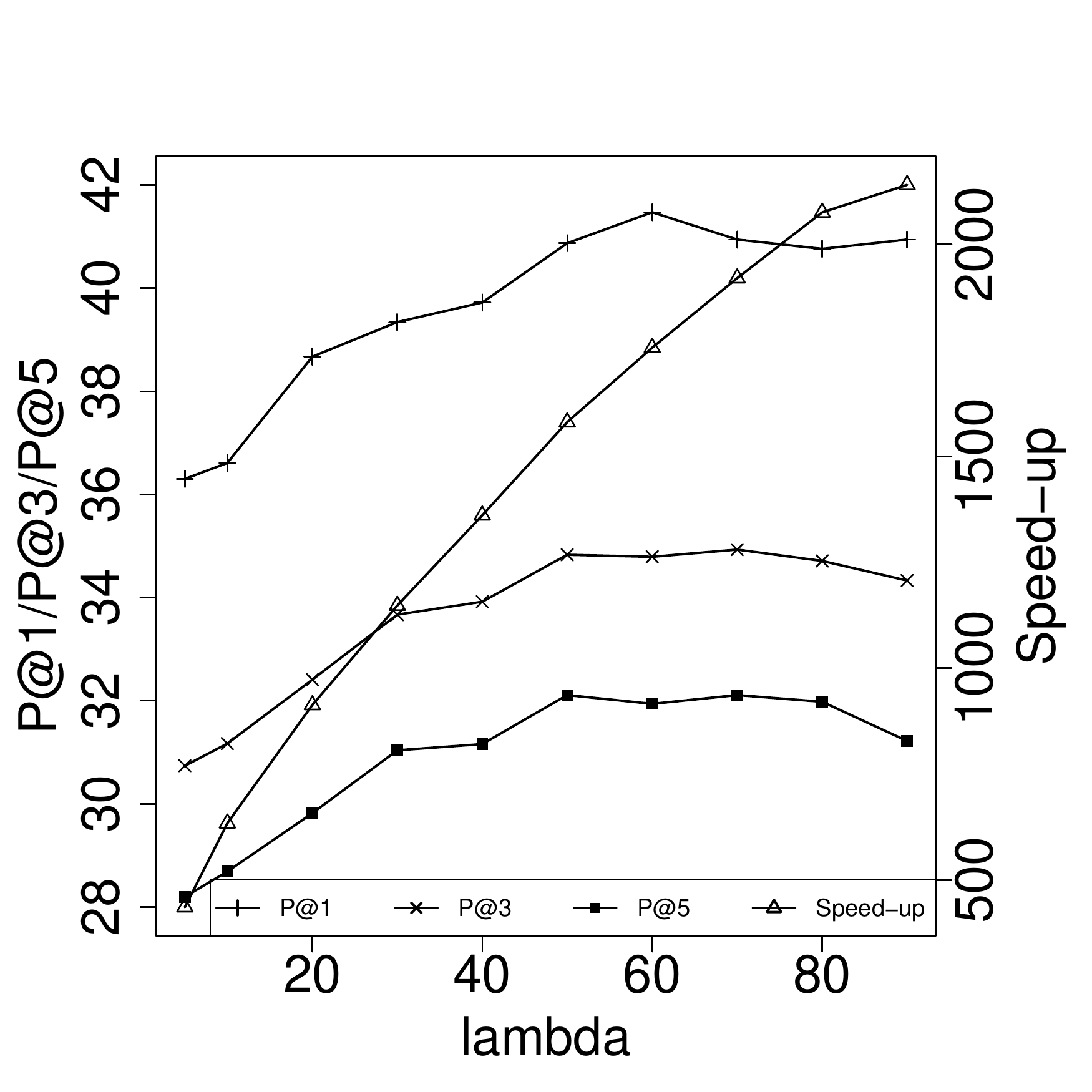}
\includegraphics[width=.33\columnwidth]{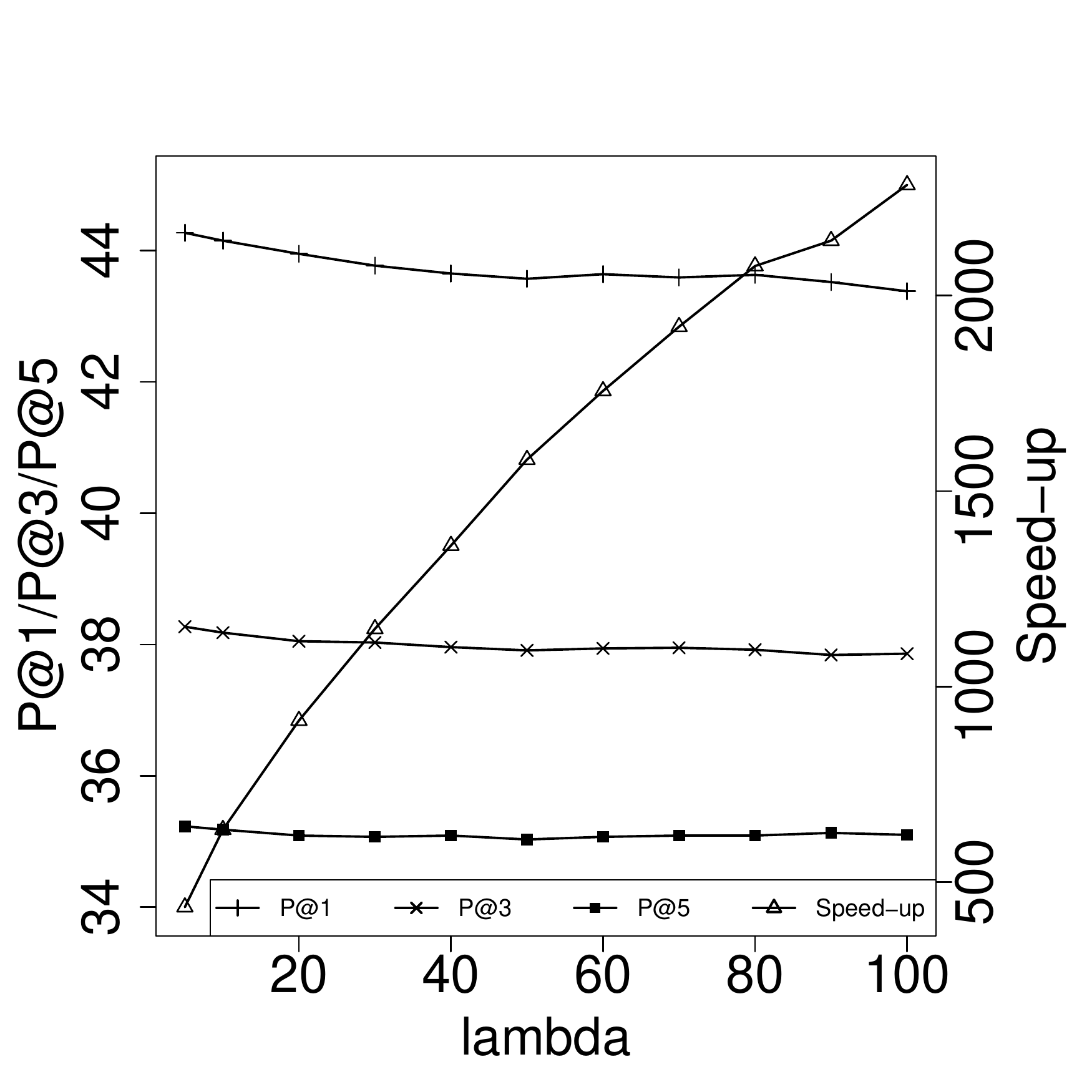}}
\caption{Precision (y1 axis) and speedup (y2 axis) as a function of $\lambda$. From \textbf{top} to \textbf{bottom}: AmazonCat-13K, Wiki10-31K, and Delicious-200K. From \textbf{left} to \textbf{right}: LEML, PD-Sparse, and DiSMEC. Presented $\lambda$ may be out of the optimal range. }
\label{resultfigure}
\end{center}
\vskip -0.3in
\end{figure*}

\textbf{Results:} We examine the performance of BP.
\begin{enumerate}
\item Prediction accuracy. Table~\ref{Ptable} demonstrates that BP speeds up prediction under all scenarios with less than $2\%$ loss in P@$k$ and less than $6\%$ loss in PSP@$k$. The accuracy-driven BP LEML produces better P@$k$ than the original LEML for all data sets, and even better PSP@$k$ in some cases. By partitioning the original label space into different subspaces, BP effectively reinforces low-rankness, a fundamental assumption that LEML must satisfy to achieve good performance. Although the original PD-Sparse does not scale to challenging data sets, BP frees it from the size constraint. 
\item Speed-up. BP speeds up prediction for up to 2780 times under all scenarios. A negative correlation is observed between speed-up and average labels per sample (ALpS) on PD-Sparse and DiSMEC. Note that ALpS in AmazonCat-13K, WikiLSHTC-325K and Amazon-670K is below $6$, and every sample in Delicious-200K has $76$ positive labels on average. Due to the nature of polished one-versus-all approach, choosing the top $5$ labels from $76$ is much easier than choosing the top $5$ from $6$. As label size in each cluster should have been substantially large to avoid unwanted elimination, speed-up may be affected. Nevertheless, even though the speed-up is less in low ALpS data sets, we still speed up PD-Sparse and DiSMEC for more than 10 times on the largest Amazon-670K.
\item Role of $\lambda$. Figure~\ref{resultfigure} reveals that $\lambda$ controls the trade-off between P@$k$ and speed-up. Even though a smaller $\lambda$ allows more labels in label clusters, minimizing $\lambda$ does not necessarily yield the best prediction performance.
\item Processing time. Table~\ref{Time} shows that our algorithm does not have much overhead in the pre-processing step. Although our goal is to speed up prediction, the training time (data partition time + BP training time) is also shorter than that of the original algorithms.
\item Comparison with CBMLC and LPSR-NB. CBMLC performs better than BP in terms of prediction accuracy with LEML, but worse with the other two, especially with PD-Sparse. The use of NB in LPSR does not allow reaching high accuracy.
\end{enumerate}

\begin{table*}[!]
\caption{P@$k$, PSP@$k$ and prediction speed-up for algorithms with BP pretreatment. The accuracy-driven BP aims at the highest P@$k$ in search space of $\lambda$. The speed-driven BP aims at the highest speed-up at the expense of at most $2\%$ P@$k$ loss. Values with superscript $\dagger$ are obtained from a subset of test samples as the built-in testing procedure of PD-Sparse does not scale to that data set.}
\label{Ptable}
\vskip -0.2in
\begin{center}
\begin{small}
\begin{tabular}{clccccccc} 
 \hline
 Data & Method & P@1 & P@3 & P@5 & PSP@1 & PSP@3 & PSP@5 & Speed-up \\
 \hline
 \multirow{13}{*}{\shortstack[c]{AmazonCat-13K\\  (ALpS=5.04)}} 
 & LEML & 88.80 & 70.65 & 54.22 & 45.62 & 52.69 & 53.41 & 1x \\
 & BP LEML (accuracy-driven) & 87.03 & 71.50 & 54.43 & 48.74 & 55.23 & 54.85 & 37x \\
 & BP LEML (speed-driven) & 86.87 & 69.94 & 52.26 & 47.72 & 53.60 & 52.32 & 58x \\
 & CBMLC LEML & 89.03 & 72.38 & 56.18 & 46.91 & 55.34 & 54.92 & 1x \\
 \cline{2-9}
 & PD-Sparse & 89.04 & 73.46 & 59.37 & 49.58 & 61.63 & 68.23 & 1x \\
 & BP PD-Sparse (accuracy-driven) & 87.38 & 71.74 & 57.40 & 48.16 & 60.44 & 67.19 & 8x \\
 & BP PD-Sparse (speed-driven) & 87.38 & 71.74 & 57.40 & 48.16 & 60.44 & 67.19 & 8x \\
 & CBMLC PD-Sparse & 87.70 & 71.34 & 57.09 & 47.33 & 59.51 & 66.40 & 1x \\
 \cline{2-9}
 & DiSMEC & 93.38 & 79.07 & 64.09 & 59.08 & 67.10 & 71.19 & 1x \\
 & BP DiSMEC (accuracy-driven) & 92.13 & 77.57 & 62.49 & 53.58 & 63.30 & 67.66 & 5x \\
 & BP DiSMEC (speed-driven) & 92.11 & 77.56 & 62.27 & 53.50 & 62.91 & 66.89 & 6x \\
 & CBMLC DiSMEC & 92.06 & 77.53 & 62.19 & 53.31 & 62.40 & 66.55 & 1x \\
 \cline{2-9}
 & LPSR-NB & 75.10 & 60.20 & 57.30 & -- & -- & -- & -- \\
 \hline
 \multirow{13}{*}{\shortstack[c]{Wiki10-31K\\ (ALpS = 18.64)}} 
 & LEML & 73.10 & 62.13 & 54.06 & 9.43 & 10.07 & 10.53 & 1x \\
 & BP LEML (accuracy-driven) & 75.57 & 62.51 & 53.73 & 9.14 & 9.48 & 9.66 & 212x \\
 & BP LEML (speed-driven) & 74.85 & 61.96 & 52.52 & 9.02 & 9.35 & 9.35 & 274x \\
 & CBMLC LEML & 74.43 & 63.32 & 55.00 & 8.97 & 10.65 & 10.93 & 1x \\
 \cline{2-9}
 & PD-Sparse & 81.89 & 65.34 & 53.74 & 12.48 & 11.89 & 12.67 & 1x \\
 & BP PD-Sparse (accuracy-driven) & 81.56 & 65.02 & 53.04 & 11.83 & 11.28 & 12.10 & 209x \\
 & BP PD-Sparse (speed-driven) & 80.89 & 64.30 & 52.62 & 11.19 & 10.56 & 11.42 & 289x \\
 & CBMLC PD-Sparse & 71.16 &  56.51 & 46.28 & 9.85 & 9.20 & 9.98 & 1x \\
 \cline{2-9}
 & DiSMEC & 85.20 & 74.90 & 65.90 & 13.55 & 13.08 & 13.81 & 1x \\
 & BP DiSMEC (accuracy-driven) & 84.08 & 74.22 & 64.95 & 10.42 & 12.07 & 12.99 & 40x \\
 & BP DiSMEC (speed-driven) & 84.08 & 73.95 & 64.30 & 10.41 & 11.90 & 12.54 & 90x \\
 & CBMLC DiSMEC & 84.01 & 73.83 & 64.22 & 10.39 & 11.79 & 12.38 & 1x \\
 \cline{2-9}
 & LPSR-NB & 72.71 & 58.51 & 49.40 & 12.79 & 12.26 & 12.13 & -- \\
 \hline
 \multirow{13}{*}{\shortstack[c]{Delicious-200K\\ (ALpS = 75.54)}} 
 & LEML & 40.01 & 36.99 & 35.07 & 6.06 & 7.24 & 8.10 & 1x \\
 & BP LEML (accuracy-driven) & 40.52 & 37.40 & 35.47 & 5.97 & 7.10 & 7.93 & 390x \\
 & BP LEML (speed-driven) & 39.67 & 37.35 & 35.44 & 5.95 & 6.89 & 7.64 & 1467x \\
 & CBMLC LEML & 42.12 & 38.05 & 35.98 & 6.84 & 7.93 & 8.88 & 1x \\
 \cline{2-9}
 & PD-Sparse & 41.97 & 35.28 & 32.87 & 5.29 & 5.80 & 6.24 & 1x \\
 & BP PD-Sparse (accuracy-driven) & 41.47 & 34.79 & 31.94 & 5.14 & 5.60 & 6.05 & 1756x \\
 & BP PD-Sparse (speed-driven) & 40.94 & 34.33 & 31.22 & 5.09 & 5.95 & 6.15 & 2104x \\
 & CBMLC PD-Sparse & 30.07 & 25.27 & 23.06 & 4.02 & 4.16 & 4.89 & 1x\\
 \cline{2-9}
 & DiSMEC & 45.53 & 38.67 & 35.52 & 6.52 & 7.61 & 8.39 & 1x \\
 & BP DiSMEC (accuracy-driven) & 44.27 & 38.27 & 35.23 & 6.66 & 7.42 & 8.03 & 436x \\
 & BP DiSMEC (speed-driven) & 43.63 & 37.92 & 35.09 & 6.53 & 7.32 & 7.96 & 2075x \\
 & CBMLC DiSMEC & 43.99 & 37.83 & 35.00 & 6.59 & 7.17 & 7.72 & 1x \\
 \cline{2-9}
 & LPSR-NB & 18.59 & 15.43 & 14.07 & 3.24 & 3.42 & 3.64 & -- \\
 \hline
 \multirow{10}{*}{\shortstack[c]{WikiLSHTC-325K\\ (ALpS = 3.19)}} 
 & LEML & 19.82 & 11.43 & 8.39 & 3.48 & 3.79 & 4.27 & 1x \\
 & BP LEML (accuracy-driven) & 29.68 & 17.86 & 13.20 & 7.30 & 8.10 & 8.97 & 84x \\
 & BP LEML (speed-driven) & 21.99 & 11.29 & 7.52 & 4.36 & 3.97 & 3.82 & 2559x \\
 \cline{2-9}
 & PD-Sparse & 60.98 & 38.12 & 26.53 & 28.34 & 33.50 & 36.62 & 1x \\
 & BP PD-Sparse (accuracy-driven) & 59.26 & 37.48 & 26.14 & 27.72 & 32.44 & 35.48 & 3x \\
 & BP PD-Sparse (speed-driven) & 59.01 & 37.33 & 26.03 & 27.63 & 32.11 & 35.39 & 4x \\
 \cline{2-9}
 & DiSMEC & 64.14 & 42.45 & 31.52 & 29.10 & 35.60 & 39.50 & 1x \\
 & BP DiSMEC (accuracy-driven) & 63.54 & 42.38 & 30.97 & 27.45 & 33.68 & 37.70 & 10x \\
 & BP DiSMEC (speed-driven) & 62.88 & 40.87 & 29.53 & 26.65 & 33.67 & 37.16 & 18x \\
 \cline{2-9}
 & LPSR-NB & 27.40 & 16.40 & 12.00 & 6.93 & 7.21 & 7.86 & -- \\
 \hline
 \multirow{10}{*}{\shortstack[c]{Amazon-670K\\ (ALpS = 5.45)}} 
 & LEML & 8.13 & 6.83 & 6.02 & 2.07 & 2.26 & 2.47 & 1x \\
 & BP LEML (accuracy-driven) & 14.21 & 10.66 & 8.60 & 5.87 & 5.47 & 5.29 & 387x \\
 & BP LEML (speed-driven) & 7.87 & 5.29 & 4.12 & 3.00 & 2.51 & 2.34 & 2780x \\
 \cline{2-9}
 & PD-Sparse & 24.41$^\dagger$ & 21.54$^\dagger$ & 19.71$^\dagger$ & 18.83$^\dagger$ & 15.77$^\dagger$ & 14.54$^\dagger$ & 1x \\
 & BP PD-Sparse (accuracy-driven) & 26.80 & 22.28 & 18.82 & 20.09 & 16.80 & 12.29 & 28x \\
 & BP PD-Sparse (speed-driven) & 23.90 & 19.27 & 17.83 & 17.87 & 14.34 & 11.78 & 42x \\
 \cline{2-9}
 & DiSMEC & 44.71 & 39.65 & 36.09 & 27.80 & 30.60 & 34.20 & 1x \\
 & BP DiSMEC (accuracy-driven) & 43.78 & 38.88 & 35.44 & 24.87 & 28.36 & 32.25 & 11x \\
 & BP DiSMEC (speed-driven) & 42.91 & 37.89 & 34.16 & 24.11 & 28.14 & 31.99 & 22x \\
 \cline{2-9}
 & LPSR-NB & 28.60 & 24.90 & 22.30 & 16.68 & 18.07 & 19.43 & -- \\
 \hline
\end{tabular}
\end{small}
\end{center}
\vskip -0.1in
\end{table*}

\section{Conclusion}

We apply BP pretreatment on extreme multi-label classifiers whose test time complexity is linear to the size of label set. The non-convex and discrete optimization problem launched with BP is solved by an intuitive and straightforward alternating minimization procedure. Partitioning on instance and label matrices embraces strength of existing algorithms by a significant order of magnitude speed-up together with a similar level of prediction accuracy.

\section{Appendix}

\subsection{LIBLINEAR Train and Prediction Time}

We report the train and prediction time, together with the number of instance clusters $q$ in all data sets for for LEML and DiSMEC in Table~\ref{LIBLINEAR}.

\begin{table*}[h]
\caption{LIBLINEAR train and prediction time (in seconds) for LEML and DiSMEC with BP pretreatment.}
\label{LIBLINEAR}
\begin{center}
\begin{small}
\begin{tabular}{clrr} 
 \hline
 Data & Method & Train & Prediction \\
 \hline
 AmazonCat-13K & BP LEML & 573 & 3 \\
 ($q=5$) & BP DiSMEC & 405 & 12 \\
 \hline
 Wiki10-31K & BP LEML & 21 & 1 \\
 ($q=3$) & BP DiSMEC & 8 & 1 \\
 \hline
 Delicious-200K & BP LEML & 1233 & 9 \\
 ($q=10$) & BP DiSMEC & 257 & 10 \\
 \hline
 WikiLSHTC-325K & BP LEML & 8447 & 8 \\
 ($q=1575$) & BP DiSMEC & 2487 & 27 \\
 \hline
 Amazon-670K & BP LEML & 3375 & 5 \\
 ($q=2000$) & BP DiSMEC & 937 & 14 \\
 \hline
\end{tabular}
\end{small}
\end{center}
\vskip -0.1in
\end{table*}

\subsection{Recall with BP Pretreatment}

As a supplement to precision, we present recall for our method versus previous method on speed-driven LEML and DiSMEC in Table~\ref{Rtable}. The definition of recall in multi-label classification is adopted from \cite{tsoumakas2007multi}. Let $n$ be the number of sample points, $Y_i$ be the ground truth label assignment of the $i$th sample, and $Z^k_i$ be the top $k$ predicted labels of the $i$th sample. Recall@$k$ (R@$k$) is defined as
\begin{equation*}
    \text{R@}k = \frac{1}{n}\sum_{i=1}^n \frac{|Y_i \cap Z^k_i|}{|Y_i|}.
\end{equation*}
\begin{table*}[h]
\caption{R@$k$ for speed-driven LEML and DiSMEC with BP pretreatment.}
\label{Rtable}
\vskip -0.2in
\begin{center}
\begin{small}
\begin{tabular}{clccc} 
 \hline
 Data & Method & R@1 & R@3 & R@5 \\
 \hline
 \multirow{4}{*}{\shortstack[c]{AmazonCat-13K\\  (ALpS=5.04)}} 
 & LEML & 24.64 & 54.03 & 64.22 \\
 & BP LEML (speed-driven) & 24.60 & 53.88 & 63.18 \\
 \cline{2-5}
 & DiSMEC & 25.18 & 56.29 & 67.99 \\
 & BP DiSMEC (speed-driven) & 25.11 & 55.97 & 67.78 \\
 \hline
 \multirow{4}{*}{\shortstack[c]{Wiki10-31K\\ (ALpS = 18.64)}}
 & LEML & 4.24 & 10.64 & 15.19 \\
 & BP LEML (speed-driven) & 4.34 & 10.60 & 14.64 \\
 \cline{2-5}
 & DiSMEC & 5.06 & 12.96 & 18.93 \\
 & BP DiSMEC (speed-driven) & 4.95 & 12.88 & 18.28 \\
 \hline
 \multirow{4}{*}{\shortstack[c]{Delicious-200K\\ (ALpS = 75.54)}} 
 & LEML & 0.87 & 2.05 & 3.07 \\
 & BP LEML (speed-driven) & 0.83 & 2.10 & 3.10 \\
 \cline{2-5}
 & DiSMEC & 1.35 & 2.60 & 3.57 \\
 & BP DiSMEC (speed-driven) & 1.13 & 2.48 & 3.50 \\
 \hline
 \multirow{4}{*}{\shortstack[c]{WikiLSHTC-325K\\ (ALpS = 3.19)}} 
 & LEML & 6.15 & 8.94 & 9.85 \\
 & BP LEML (speed-driven) & 6.28 & 8.84 & 9.50 \\
 \cline{2-5}
 & DiSMEC & 19.94 & 33.23 & 37.01 \\
 & BP DiSMEC (speed-driven) & 19.07 & 32.55 & 36.43 \\
 \hline
 \multirow{4}{*}{\shortstack[c]{Amazon-670K\\ (ALpS = 5.45)}}
 & LEML & 1.81 & 3.37 & 4.05 \\
 & BP LEML (speed-driven) & 1.67 & 3.13 & 3.93 \\
 \cline{2-5}
 & DiSMEC & 9.94 & 19.57 & 24.28 \\
 & BP DiSMEC (speed-driven) & 9.63 & 19.21 & 24.09 \\
 \hline
\end{tabular}
\end{small}
\end{center}
\vskip -0.1in
\end{table*}

\end{document}